\pdfoutput=1

\documentclass[11pt]{article}

\usepackage{ACL2023}

\usepackage{times}
\usepackage{latexsym}
\usepackage{graphicx}
\usepackage{longtable}
\usepackage{multirow, makecell}
\usepackage{booktabs}
\usepackage{soul}
\usepackage{bbm}
\usepackage{hyperref}
\usepackage{amsmath}
\usepackage{amsthm}
\usepackage{amsfonts}
\usepackage{comment}
\usepackage{subcaption}

\usepackage[T1]{fontenc}

\usepackage[utf8]{inputenc}

\usepackage{microtype}

\usepackage{inconsolata}

\newif\ifcomments
\commentstrue
\ifcomments
    \newcommand\sanjay[1]{{\color{blue}\{\textit{#1}\}$_{sanjay}$}}
    \newcommand\trevor[1]{{\color{pink}\{\textit{#1}\}$_\text{trevor}$}}
    \newcommand{\medhini}[1]{\textcolor{red}{Medhini:\ #1}}
    
\else
    \newcommand\sanjay[1]{}
    \newcommand\trevor[1]{}
    \newcommand\medhini[1]{}
\fi

\usepackage{inconsolata}
\usepackage[T1]{fontenc}

\definecolor{light-gray}{rgb}{0.8, 0.8, 0.8}
\definecolor{prompt-gray}{HTML}{a7a7a7}
\definecolor{comment-green}{rgb}{0.435, 0.576, 0.106}
\definecolor{highlight}{HTML}{e3eeff}
\definecolor{code-function}{HTML}{693da8}
\definecolor{code-syntax}{HTML}{0060b1}
\definecolor{code-constant}{HTML}{d86001}
\definecolor{tab-blue}{rgb}{0.12,0.46,0.70}
\definecolor{tab-orange}{rgb}{1,0.5,0.054}
\renewcommand\fbox{\fcolorbox{light-gray}{white}}
\newcommand{\hlcode}[1]{\colorbox{highlight}{\makebox[0.99\linewidth][l]{#1}}}
\newcommand{\query}[1]{\textcolor{comment-green}{#1}}

\newcommand{\ie}{i.e., }
\newcommand{\eg}{e.g., }

\title{Modular Visual Question Answering via Code Generation}

\author{\makecell{Sanjay Subramanian$^{1}$ ~~~~~~~ Medhini Narasimhan$^{1}$ ~~~~~~~ Kushal Khangaonkar$^{1}$ \\ Kevin Yang$^{1}$ ~~~~~ Arsha Nagrani$^{2}$ ~~~~~ Cordelia Schmid$^{2}$ ~~~~~ Andy Zeng$^{2}$ \\ Trevor Darrell$^{1}$ ~~~~~ Dan Klein$^{1}$} \\ 
$^{1}$UC Berkeley\hspace{5mm}
$^{2}$Google Research\hspace{5mm} \\
\texttt{\makecell{\{sanjayss,medhini,kushaltk,yangk,trevordarrell,klein\}@berkeley.edu, \\\{anagrani,cordelias,andyzeng\}@google.com}}}

\begin{document}
\maketitle


\begin{abstract}
\looseness=-1
We present a framework that formulates visual question answering as modular code generation. In contrast to prior work on modular approaches to VQA, our approach requires no additional training and relies on pre-trained language models (LMs), visual models pre-trained on image-caption pairs, and fifty VQA examples used for in-context learning. The generated Python programs invoke and compose the outputs of the visual models using arithmetic and conditional logic. Our approach improves accuracy on the COVR dataset by at least 3\% and on the GQA dataset by roughly 2\% compared to the few-shot baseline that does not employ code generation.
\end{abstract}

\section{Introduction}
The scope of reasoning needed for visual question answering (VQA) is vast, and demands the synthesis of many skills – from grounding language to pixels \citep{vqav2,radford2021learning,zhai2022lit} and spatial reasoning \citep{gqa} to commonsense and knowledge-based reasoning \citep{okvqa}. Consider the question \emph{``Is the carriage to the right of a horse?''}. To consistently answer such questions correctly, a system must recognize that the question is the conjunction of two subquestions: \emph{``Is there a horse?''} and \emph{``Is the carriage to the right of the horse?''} Scaling the typical finetuning paradigm to all possible combinations of reasoning skills is prohibitively expensive in annotation cost and makes it difficult to add skills to an already-trained system.

Modular approaches, on the other hand -- from classic methods \citep{krishnamurthy2013jointly}, to differentiable neural module networks (NMNs) \citep{nmn, nmn2,Saqur2020MultimodalGN}) -- offer a potential route to leverage and scale to the compositional nature of visual reasoning as a means to generalize: \ie \textit{infinite use of finite means}. However, the modules of an NMN must still be trained jointly on a large dataset, and are also restricted in that they (i) require a parser, which must be modified if modules are added or removed from the system, and (ii) require retraining if a module is replaced.

\begin{figure*}
    \centering
    \includegraphics[width=\textwidth]{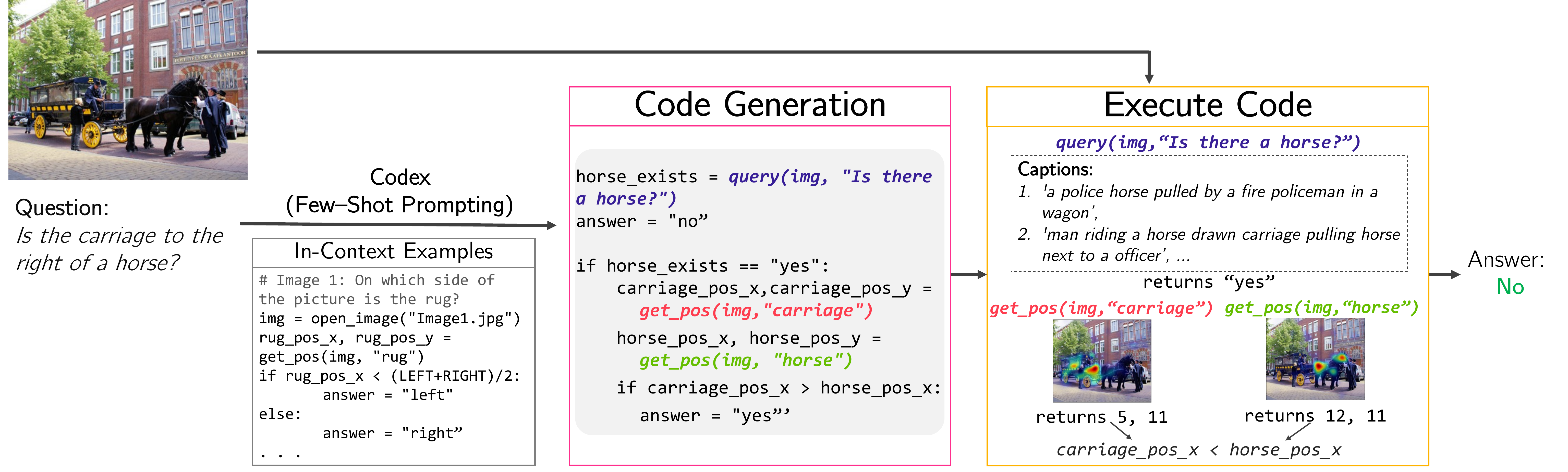}
    \caption{\small{\textbf{\texttt{CodeVQA Overview.}} \texttt{CodeVQA} first prompts Codex with in-context examples that break down a given question into Python code. Using just the question, Codex generates an executable program that composes pre-defined visual modules using conditional logic, arithmetic, etc. The visual module, \texttt{query} answers a question by  captioning the image and using an LM to answer based on the captions. \texttt{get\_pos} retrieves the location of the object. Here, CodeVQA correctly identifies the question as a conjunction of a query and a spatial comparison and arrives at the right answer.    
    }}
    \label{fig:overview}
\end{figure*}

In this work, we investigate an alternative class of modular VQA approaches, whereby building on the recent advent of highly capable out-of-the-box language models (LMs) \cite{codex,Ouyang2022TrainingLM} and visual language models (VLMs) \cite{blip}, we develop systems that formulate VQA as a program synthesis problem. Specifically, our method \texttt{CodeVQA}, illustrated in Figure~\ref{fig:overview}, uses code-writing LMs to take questions as input, and outputs code to (i) orchestrate a series of visual primitive APIs that wrap around VLMs to probe the image for specific pieces of visual information (\eg captions, pixel locations of entities, or image-text similarity scores), and (ii) reason about that information with the full expression of Python code (e.g. arithmetic, logic structures, feedback loops, etc.) to arrive at an answer. From a practical perspective, the modularity of \texttt{CodeVQA} combined with the few-shot prompting capabilities of LMs enable it to adapt to a broad range of desired VQA label distributions without additional model training, and benefits from replacing individual modules with improved versions as they become available.

We evaluate \texttt{CodeVQA} in the few-shot VQA setting, which has seen a great deal of recent work \citep{flamingo,fewvlm,pica,pnpvqa}. Our method outperforms previous approaches by at least 3\% on the COVR dataset \citep{covr}, which requires reasoning over multiple images, and by roughly 2\% on the GQA dataset \citep{gqa}. Our results suggest that the benefits of modularity with recent off-the-shelf models can be realized in VQA without additional model training.\footnote{Our code and annotated programs are available at \url{https://github.com/sanjayss34/codevqa}.}
\section{Related Work}
\label{sec:related}
\noindent
Several recent approaches for reasoning tasks consist of an LM that writes programs and an interpreter for these programs. \citet{Liang2022CodeAP} applies this approach to robotics. \citet{binder} introduces a framework for reasoning jointly over tables, text, and images, where the images are represented by image captions. \citet{subramanian-etal-2022-reclip} used a syntactic parser and hard-coded rules rather than an LM to aggregate outputs from CLIP \citep{radford2021learning} for zero-shot referring expression comprehension; their finding that CLIP is not useful for spatial keywords motivates our code generation approach to spatial reasoning.

\noindent
Concurrent with our work, other papers have introduced similar frameworks for multi-hop VQA \citep{visprog,vipergpt}. These papers conflate the benefit of program synthesis with the benefits of the LM, in-context examples, and vision models used as primitives. By contrast, we analyze the effect of program synthesis by comparing \texttt{CodeVQA} against a strong LM-based few-shot baseline using the same in-context example selection method. Moreover, while these frameworks rely on supervised VQA or object detection models, we show that we can obtain comparable performance (on the GQA dataset) using only the LM and models pre-trained on image-text pairs.
\section{Few-shot VQA via Code Generation}
\label{sec:method}
In visual question answering (VQA), the inputs to the system are an image and a question and the output is a textual answer. We consider the few-shot VQA setting in which the system has access to only a small number (50) of human-annotated VQA instances.

\noindent
\textbf{Overview.} Fig~\ref{fig:overview} illustrates our approach. Given an image and a corresponding question, \texttt{CodeVQA} first generates a Python program using just the question. It then executes this program, using the image when necessary, to predict the answer. We first define the set of code primitives that our system uses (\S~\ref{sec:primitives}). Then we describe how we generate a program that composes these primitives based on the question (\S~\ref{sec:code_gen}). Finally, we enumerate the pre-trained models that we employ (\S~\ref{sec:component_models}).

\subsection{Code Primitives}
\label{sec:primitives}
Primitives define basic operations over the image or over text that are often useful for VQA. In \texttt{CodeVQA}, we use three primitives, which are defined below. Each of these primitives is implemented using image-text matching (ITM), image-text contrastive (ITC), and image-captioning models, each of which can be trained with only image-caption pairs.
The difference between ITM and ITC is that ITC computes separate image and text embeddings and takes a dot product, while ITM performs early fusion on the image and text features and is thus more computationally expensive. We note that our framework is not tied to this choice of primitives and can support other, more complex primitives that could draw on other aspects of the programming language and third-party libraries.
\vspace{-1mm}
\paragraph{\texttt{query(image, question)}} This function answers a question about the given image. Our implementation of this function is based on PnP-VQA \citep{pnpvqa} and PICa \citep{pica} and is implemented with the following steps: (1) using the ITM model, compute the GradCAM \citep{gradcam} between the question and the image (averaged over question tokens), (2) sample $K=20$ image patches based on their GradCAM score, (3) generate a captions from the sampled patches using the captioning model, (4) Repeat steps (2) and (3) until $C$ unique captions have been generated, and (5) predict the answer by prompting an LM with the question, captions, and in-context examples. The in-context examples in step (5) are selected as described in \S~\ref{sec:code_gen}. When the dataset involves reasoning over multiple images, each in-context example has the captions for all images.

\paragraph{\texttt{get\_pos(image, text)}} This function computes the GradCAM between the given text tokens and the image using the ITM model and returns the (x, y) pair that maximizes the GradCAM value. Note that this application of GradCAM is different from the one in \texttt{query} since we do not average over all question tokens. See Appendix~\ref{app:gradcam} for more information on how we compute GradCAM maps.

\paragraph{\texttt{find\_matching\_image(images, text)}} In the setting where multiple images are associated with each question, there are questions that refer specifically to one image (e.g.\ ``What is the woman holding?''). This function can be used to select the most relevant image from the set. It is implemented by scoring each image with the text using the ITC model and picking the image with the highest score.

\subsection{Code generation}
\label{sec:code_gen}
In the first stage of \texttt{CodeVQA}, we generate a Python program based on the question. Using Python over a domain-specific language is advantageous because (1) it supports arithmetic as well as control flow including loops and if statements \citep{Liang2022CodeAP}--all of which we use in our programs--and (2) large LMs for code generation (e.g.\ Codex \citep{codex}) have been trained on a large amount of Python code.

We construct a prompt that consists of an instruction, constants that define the dimensions of the image, and import statements and API documentation (as a code comment) that specify the available functions. In addition to the prompt, the input to the LM also includes expert-annotated programs for several in-context examples. An in-context example for few-shot prompting on the COVR dataset is shown below~(question in {\color{prompt-gray}gray}, the program is \colorbox{highlight}{highlighted}).
\vspace{0.3em}
\noindent\fbox{\parbox{0.97\linewidth}{\small{\texttt{{
{\color{prompt-gray}\# Image Set 1: How many images contain exactly 2 pink shoes??}\\
\hlcode{images = {\color{code-function}open\_images}("ImageSet1.jpg")}\\
\hlcode{count = 0}\\
\hlcode{for image in images:}
\hlcode{\hspace*{8mm}two\_pink\_shoes = {\color{code-function}query}(image, "Are}\\
\hlcode{\hspace*{16mm}there exactly 2 pink shoes?")}\\
\hlcode{\hspace*{8mm}if two\_pink\_shoes == "yes":}\\
\hlcode{\hspace*{16mm}count += 1}\\
\hlcode{answer = count}\\
}}}}}\\
For an example of the rest of the prompt for the LM, see Appendix~\ref{app:prompt_examples}. When executing the generated program results in a runtime error, we return call \texttt{query} on the image and the original question (including captions for all images if the instance involves multiple images).

Since all annotated programs cannot fit into a single input to the model, we must select which programs to use as in-context examples for each test question. Following \citet{vidil}, we use sentence embeddings\footnote{\href{https://huggingface.co/sentence-transformers/all-mpnet-base-v2}{https://huggingface.co/sentence-transformers/all-mpnet-base-v2}} to retrieve the most similar questions for each test question.
\subsection{Component models}
\label{sec:component_models}
Our approach relies on four pre-trained models: a code generation model, an ITM model, an ITC model, an IC model, and a question-answering LM for answering questions based on captions. We use the \texttt{code-davinci-002} model \citep{codex} via the OpenAI API for both generating programs and for question-answering.
We use the BLIP models \citep{blip} finetuned for ITM, ITC, and captioning.
\begin{table}[t]
\large
\centering
\resizebox{0.45\textwidth}{!}{
\begin{tabular}{l|c|c|c}
\toprule
\multirow{2}[3]{*}{{\bf Model}} & \multicolumn{1}{c}{GQA} & \multicolumn{1}{c}{COVR} & \multicolumn{1}{c}{NLVR2} \\
& \multicolumn{1}{c}{Acc.} &  \multicolumn{1}{c}{Acc.} &  \multicolumn{1}{c}{Acc.} \\
\midrule
\textbf{Finetuned} & & &\\
VisualBERT & -- & 57.9 & 67.0 \\
VinVL-Base & 65.1 & -- & 83.1 \\
\midrule
\textbf{Zero-shot} & & & \\
FewVLM & 29.3 & -- & -- \\
PnP-VQA & 42.3 & -- & -- \\
BLIP-v2* & 44.7 & -- & -- \\
\midrule
\textbf{Few-shot} & & &  \\
FewVLM & 35.7 & -- & -- \\
VisProg* & 50.5\textdagger & -- & 62.4 \\
ViperGPT* & 48.2 & -- & -- \\
Few-shot PnP-VQA & 46.6 & 45.8 & 63.4 \\
\texttt{CodeVQA} (ours) & \textbf{49.0} & \textbf{50.7} & \textbf{64.0} \\
\bottomrule
\end{tabular}
}
\caption{\textbf{Results on GQA (testdev), COVR (test), and NLVR2 (test-public)} datasets from \texttt{CodeVQA}, Few-shot PnP-VQA, prior work VisualBERT \citep{visualbert}, VinVL-Base \citep{vinvl}, FewVLM \citep{fewvlm}, PnP-VQA \citep{pnpvqa}, and concurrent work (marked with *) BLIP-v2 \citep{blipv2}, VisProg \citep{visprog}, and ViperGPT \citep{vipergpt}. Our method outperforms all few-shot methods from prior work. Highest few-shot scores for each full dataset are in \textbf{bold}. \textdagger indicates an evaluation on a stratified sample of the test set, which may not be directly comparable to results on the full test set.}
\label{tab:main-results}
\vspace{-6mm}
\end{table}
\section{Experiments}

\subsection{Implementation Details}
See Appendix~\ref{app:implementation} for implementation details.
\subsection{Datasets}
The GQA dataset \citep{gqa} contains multi-hop questions generated from human-annotated scene graphs of individual images in Visual Genome \citep{visualgenome}. The COVR dataset \citep{covr} contains multi-hop questions about \emph{sets of images} in the Visual Genome and imSitu \citep{imsitu} datasets. These questions are synthetically generated from templates and are then paraphrased by humans. Unless otherwise specified, we present results on the paraphrased questions. The NLVR2 dataset \citep{nlvr2} contains statements about \emph{pairs of images}, and the task is to determine whether each statement is true or false (we rephrase the statements as questions before feeding it to the methods that we evaluate). Appendix~\ref{app:datasets} has further details about the datasets.
For each of the three datasets, we wrote programs for 50 questions randomly sampled from the corresponding training set. Unless stated otherwise, we put 12 in-context examples in a prompt for a single-image dataset and 6 in-context examples in a prompt for a multi-image dataset (since including captions for multiple images increases the necessary context size for each example). We report the exact-match accuracies of the lower-cased answers.

\subsection{Baseline}
Our primary baseline is an adaptation of PnP-VQA \citep{pnpvqa} to the few-shot setting. We refer to it as ``Few-shot PnP-VQA.'' This baseline is equivalent to running the five-step \texttt{query} procedure described in \S~\ref{sec:primitives} for every question. We also compare to zero-shot and few-shot methods from prior and concurrent work. Appendix~\ref{app:baselines} contains further details about those methods.

\subsection{Results}
Table~\ref{tab:main-results} shows the results on the three datasets. \texttt{CodeVQA} has the highest accuracy among the few-shot techniques. Compared to Few-shot PnP-VQA, it has markedly better performance on COVR, which makes sense because in this dataset, the baseline approach must combine information across image captions for multiple images when given a single prompt. On the other hand, our method loops over the images and queries a single image at a time or selects the image most relevant to the question. Indeed, Table~\ref{tab:covr-num-images} shows that \texttt{CodeVQA} has the greatest advantage on instances involving 4 or 5 images.
Compared to concurrent work that also uses program synthesis, \texttt{CodeVQA} generally performs better, which could be due to methodological differences like our in-context example retrieval or due to implementation details.

\begin{figure*}[ht]
    \centering
    \includegraphics[width=\textwidth]{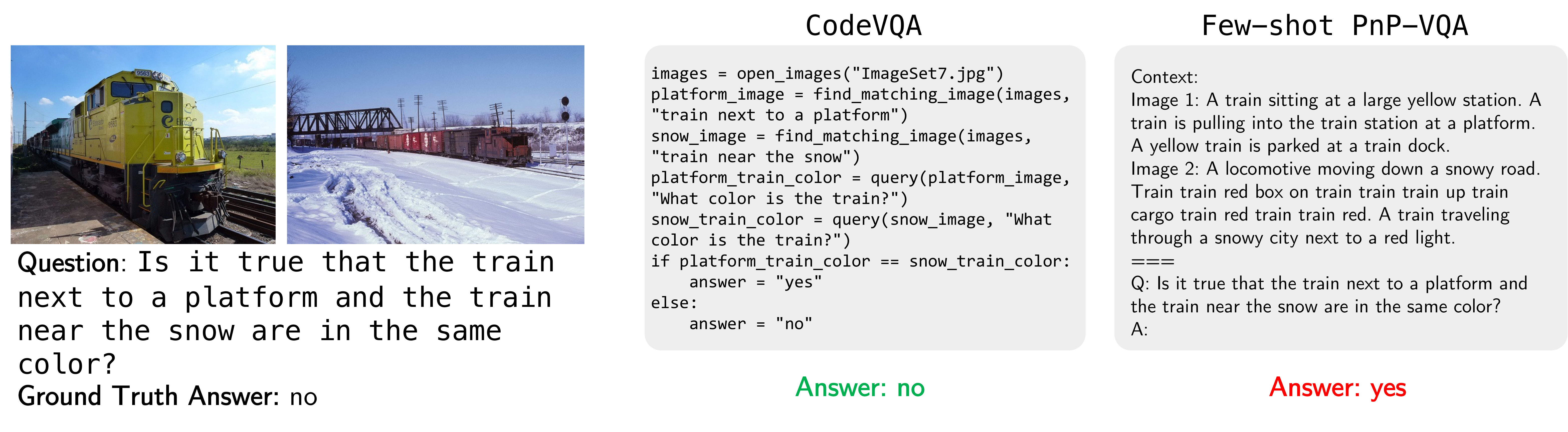}
    \vspace{-10mm}
    \caption{\small{\textbf{Qualitative Comparison}. \texttt{CodeVQA} correctly answers this question from COVR by breaking it into simpler questions, answering each separately, and comparing the answers. Few-shot PnP-VQA answers incorrectly, even though the captions contain the necessary information. (Note that \texttt{CodeVQA} also generates captions, which are not shown here.)}    
    }
    \label{fig:res1}
\end{figure*}

Fig.~\ref{fig:res1} shows a qualitative comparison of \texttt{CodeVQA} and the baseline Few-shot PnP-VQA on the COVR dataset. \texttt{CodeVQA} answers the question correctly by answering a simpler question for each image and comparing the answers, while Few-shot PnP-VQA answers incorrectly despite producing captions with the necessary information.

\subsection{Ablations}
Table~\ref{tab:example-selection} compares embedding-based retrieval of in-context examples with random retrieval. \texttt{CodeVQA}'s improvement over Few-shot PnP-VQA is greater when in-context examples are retrieved by embedding. Embedding-based retrieval offers a systematic way to collect relevant in-context examples rather than curating a single set of examples as in \citet{visprog}.

In Appendix~\ref{app:quant-results}, we include ablations for the question-answering LM and for the number of shots in the prompt as well as results on validation sets. Table~\ref{tab:results-with-val} shows that \texttt{CodeVQA} improves over Few-shot PnP-VQA when either \texttt{code-davinci-002} or \texttt{text-davinci-003} is used as the question-answering LM. Table~\ref{tab:shot-number} shows roughly constant accuracy as the number of in-context examples is varied.

\begin{table}[ht]
\small
\centering
\captionsetup{font=footnotesize}
\resizebox{0.5\textwidth}{!}{
\begin{tabular}{lcc}
\toprule
Retrieval Method & Few-shot PnP-VQA & \texttt{CodeVQA} \\
\midrule
\textit{text-davinci-003} & & \\
Random & 48.15 & 49.9 \\
Embedding & 49.4 & 52.5 \\
\midrule
\textit{code-davinci-002} & & \\
Random & 49.5 & 50.7 \\
Embedding & 52.1 & 55.3 \\
\bottomrule
\end{tabular}
}
\caption{\textbf{Comparing Example Retrieval Techniques} on 2000 GQA validation examples. Italicized GPT model name denotes the model used as the question-answering LM.}
\label{tab:example-selection}
\vspace{-3mm}
\end{table}

\subsection{Analysis}
\label{sec:analysis}
\begin{figure}
    \vspace{-2mm}
    \includegraphics[width=0.48\textwidth]{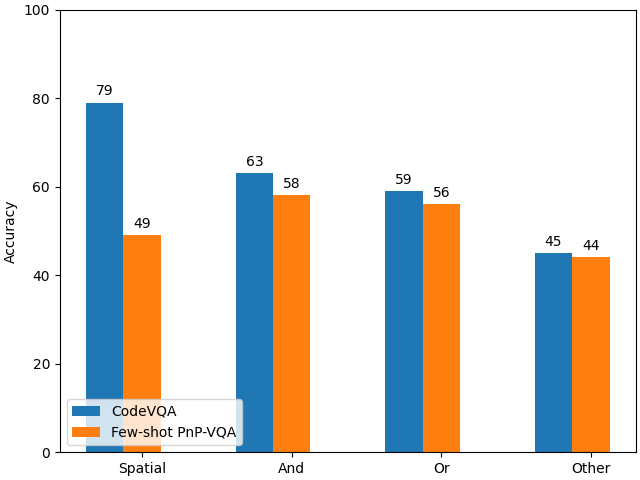}
    \caption{\textbf{Accuracy by question type in GQA test set.} \textcolor{tab-blue}{\texttt{CodeVQA} (blue)} outperforms \textcolor{tab-orange}{Few-shot PnP-VQA (orange)} on the spatial, and, or questions. ``Spatial'' refers to questions focusing on left/right or top/bottom relations or object positions.}
    \label{fig:gqa_breakdown2}
    \vspace{-5mm}
\end{figure}

Figure~\ref{fig:gqa_breakdown2} breaks down accuracy
by question type. \texttt{CodeVQA}'s greatest improvement (roughly 30\%) is in the subset consisting of questions about left/right or top/bottom object positions. There is also an improvement in ``and'' and ``or'' questions. This improvement could be related to the recent finding that LMs benefit from converting multi-hop into single-hop questions \citep{selfask}.\footnote{Accuracy on this kind of question can also be improved by improving the LM. For instance, using \texttt{text-davinci-003} as the LM for QA closes the gap between Few-shot PnP-VQA and \texttt{CodeVQA} on ``and'' questions in GQA.}

\begin{table}[ht]
\vspace{-2mm}
\small
\centering
\captionsetup{font=footnotesize}
\resizebox{0.5\textwidth}{!}{
\begin{tabular}{lccccc}
\toprule
& \multicolumn{5}{c}{Number of images} \\
 & 1 & 2 & 3 & 4 & 5 \\
 \midrule
 \# of Instances & 12 & 915 & 828 & 696 & 4440 \\
Few-shot PnP-VQA & 91.7 & 51.5 & 48.3 & 47.0 & 46.9  \\
\texttt{CodeVQA} & 75.0 & 53.3 & 48.7 & 53.2 & 53.4 \\
\bottomrule
\end{tabular}
}
\caption{\textbf{Accuracy by number of images per instance} on COVR validation set.}
\vspace{-3mm}
\label{tab:covr-num-images}
\end{table}

We analyzed sources of error in \texttt{CodeVQA} on 100 examples in the COVR validation set for which \texttt{CodeVQA} answered incorrectly: irrelevant captions (31\%), mistake in \texttt{find\_matching\_image} (12\%), program generation error (14\%), question-answering error (25\%), predicted answer could be considered correct (14\%), ground-truth is unclear/incorrect (16\%), and numerical error (1\%). Note that these categories are not mutually exclusive, and 13 of the 100 examples were marked with multiple categories. Thus, more errors are due to execution of the modules than program generation. 

\vspace{-6mm}
\section{Conclusion}

\label{sec:conclusion}
In this paper, we have introduced a framework for modular few-shot VQA. Our approach prompts an LM to generate a Python program that invokes pre-trained visual modules and composes the outputs of these modules to predict the answer. Unlike previous modular VQA techniques, this framework does not require (re-)training modules or a parser. Also, obtaining interpretable module outputs from previous modular approaches is nontrivial \citep{Subramanian2020ObtainingFI}, whereas in our approach the modules are frozen and thus interpretable. \texttt{CodeVQA} can also be viewed as expanding pipelined systems \citep{zeng2022socratic} to the full expression of code. Our approach exhibits  empirical gains, motivating future work on modular few-shot VQA.
\vspace{-2mm}
\section{Limitations}
While the initial results are promising, the accuracy of our method remains lower than human VQA accuracy and models finetuned on the VQA datasets, which suggests that there may still be substantial progress that must be made before few-shot VQA methods with code synthesis are useful for practical real world applications. Also, further work is needed on extending the framework to additional primitives, as the results in Appendix~\ref{app:additional} show that doing so does not always lead to improvements over the baseline method.
Another limitation of our approach is that it relies on large capable LMs, which may be restricted in use due to compute requirements or cost (e.g. via available APIs). We also focus in this work on benchmarking VQA capabilities with English as the primary language – future work may extend this to other languages via multilingual LMs.

\section{Acknowledgements}
We thank the members of the Berkeley NLP group (especially Eric Wallace), Grace Luo, and the anonymous reviewers for feedback on earlier drafts of this paper. We are grateful to Ben Bogin and Shivanshu Gupta for their assistance in evaluating \texttt{CodeVQA} and Few-shot PnP-VQA on the private COVR test set. SS, MN, and TD were supported in part by the DoD, including an NDSEG fellowship (for SS) and DARPA’s LwLL, PTG, and/or SemaFor programs, by the NSF, and/or by the Berkeley Artificial Intelligence Research (BAIR) industrial alliance program.

\bibliography{anthology,custom}
\bibliographystyle{acl_natbib}

\appendix
\section{Code generation prompts}
\label{app:prompt_examples}
\subsection{GQA}
The preamble of the prompt ({\color{prompt-gray}gray})--containing the instruction, constants, import statements, and API documentation--and a single in-context example are shown below (question in \query{green}, program \colorbox{highlight}{highlighted}). In our main GQA experiments, 12 in-context examples are used for each evaluation example.
\vspace{0.3em}
\noindent\fbox{\parbox{0.97\linewidth}{\small{\texttt{{
{\color{prompt-gray}"""Write Python code to answer the questions about each image."""}\\
{\color{prompt-gray}\# Global constants}\\
{\color{prompt-gray}\# min x coordinate}
{\color{prompt-gray}LEFT = 0}\\
{\color{prompt-gray}\# min y coordinate}
{\color{prompt-gray}BOTTOM = 0}
{\color{prompt-gray}\# max x coordinate}
{\color{prompt-gray}RIGHT = 24}
{\color{prompt-gray}\# max y coordinate}
{\color{prompt-gray}TOP = 24}
{\color{prompt-gray}from PIL import Image}
{\color{prompt-gray}from utils import open\_images, query, find\_matching\_image, get\_pos}\\
{\color{prompt-gray}"""}\\
{\color{prompt-gray}API Reference:}\\
{\color{prompt-gray}open\_image(path: str) -> Image - opens the image at the path and returns it as an Image object}\\
{\color{prompt-gray}query(img: Image, question: str) -> str - queries the image returns an answer to the question}\\
{\color{prompt-gray}get\_pos(img: Image, object: str) -> (float, float) - returns the position of the object in the image}
{\color{prompt-gray}"""}\\
\query{\# Image 1: Does the bench look silver and metallic?}\\
\hlcode{img = {\color{code-function}open\_image}("Image1.jpg")}\\
\hlcode{is\_silver = {\color{code-function}query}(img, "Does the bench look}\\
\hlcode{\hspace*{4mm}silver and metallic?")}\\
\hlcode{is\_metallic = {\color{code-function}query}(img, "Does the bench look}\\
\hlcode{\hspace*{4mm}metallic?")}\\
\hlcode{if is\_silver == "yes" and is\_metallic == "yes":}\\
\hlcode{\hspace*{8mm}answer = "yes"}\\
\hlcode{else:}\\
\hlcode{\hspace*{8mm}answer = "no"}\\
}}}}}
\vspace{90mm}
\subsection{COVR}
The preamble of the prompt ({\color{prompt-gray}gray})--containing the instruction, constants, import statements, and API documentation--and a single in-context example (question in \query{green}, program \colorbox{highlight}{highlighted}) are shown below. In our COVR experiments, 6 in-context examples are used for each evaluation example.
\vspace{0.3em}
\noindent\fbox{\parbox{0.97\linewidth}{\small{\texttt{{
{\color{prompt-gray}"""Write Python code to answer the questions about each image."""}\\
{\color{prompt-gray}\# Global constants}\\
{\color{prompt-gray}\# min x coordinate}
{\color{prompt-gray}LEFT = 0}\\
{\color{prompt-gray}\# min y coordinate}
{\color{prompt-gray}BOTTOM = 0}
{\color{prompt-gray}\# max x coordinate}
{\color{prompt-gray}RIGHT = 24}
{\color{prompt-gray}\# max y coordinate}
{\color{prompt-gray}TOP = 24}
{\color{prompt-gray}from PIL import Image}
{\color{prompt-gray}from utils import open\_images, query, find\_matching\_image, get\_pos}\\
{\color{prompt-gray}"""}\\
{\color{prompt-gray}API Reference:}\\
{\color{prompt-gray}open\_image(path: str) -> List[Image] - opens the images in the given directory and returns them in a list of Image objects}\\
{\color{prompt-gray}query(img: Image, question: str) -> str - queries the region of the image in the given coordinates and returns an answer}\\
{\color{prompt-gray}find\_matching\_image(images: List[Image], text: str) -> Image - returns the image that best matches the text}\\
{\color{prompt-gray}get\_pos(img: Image, object: str) -> (float, float) - returns the position of the object in the image}
{\color{prompt-gray}"""}\\
\query{\# Image Set 1: Is it true that there are more ladies that are wearing black shirt than men that are wearing black shirt?}\\
\hlcode{images = {\color{code-function}open\_images}("ImageSet1.jpg")}\\
\hlcode{ladies\_total = 0}\\
\hlcode{men\_total = 0}\\
\hlcode{for image in images:}\\
\hlcode{\hspace*{8mm}ladies\_exist = {\color{code-function}query}(image, "Is there a}
\hlcode{\hspace*{12mm}lady?")}\\
\hlcode{\hspace*{8mm}if ladies\_exist == "yes":}\\
\hlcode{\hspace*{16mm}ladies\_count = {\color{code-function}int}({\color{code-function}query}(image, "How}\\
\hlcode{\hspace*{20mm}many ladies are wearing black}\\
\hlcode{\hspace*{20mm}shirt?"))}\\
\hlcode{\hspace*{16mm}ladies\_total += ladies\_count}\\
\hlcode{\hspace*{8mm}man\_exist = {\color{code-function}query}(image, "Is there a}\\
\hlcode{\hspace*{12mm}man?")}\\
\hlcode{\hspace*{8mm}if men\_exist == "yes":}\\
\hlcode{\hspace*{16mm}men\_count = {\color{code-function}int}({\color{code-function}query}(image, "How}\\
\hlcode{\hspace*{20mm}many men are wearing black}\\
\hlcode{\hspace*{20mm}shirt?"))}\\
\hlcode{\hspace*{16mm}men\_total += men\_count}\\
\hlcode{if ladies\_total > men\_total:}\\
\hlcode{\hspace*{8mm}answer = "yes"}\\
\hlcode{else:}\\
\hlcode{\hspace*{8mm}answer = "no"}\\
}}}}}\\

\section{GradCAM}
\label{app:gradcam}
Our computation of GradCAM follows prior work that uses vision transformers \citep{pnpvqa,albef}. We are given a question with tokens $q_1, ..., q_T$ and an image that is tokenized into $K \times K$ patches. We use layer $L=6$ to compute GradCAM, following \citet{pnpvqa}. We compute a GradCAM map for each token as follows. Let $C \in \mathbb{R}^{T \times K^2}$ be the cross-attention map from layer $L$. Let $G \in \mathbb{R}^{T \times K^2}$ be the gradient of the image-text matching score with respect to $C$. Then the GradCAM map for token $i$ is given by the $i$th row of $C \bigodot ReLU(G))$, where $\bigodot$ denotes elementwise multiplication. As stated in Section~\ref{sec:primitives}, for the \texttt{query} primitive, we take the average GradCAM map across all question tokens, whereas for the \texttt{get\_pos} primitive, we take the average GradCAM map across the input text tokens (which are part of the question tokens).

\section{Implementation Details}
\label{app:implementation}
To generate captions for in-context examples in each dataset, we run steps $1-4$ for each of the 50 questions in the database of in-context examples. For GQA experiments, we use $C=7$ captions per image, and for COVR experiments, where each question is associated with multiple images, we use $C=3$ captions per image.\footnote{We chose this number of captions to be the maximum possible subject to the number of shots and the context size of the \texttt{davinci} model, which we used as our question-answering LM in preliminary experiments.} We use $C=7$ captions for the NLVR2 dataset. Each reported accuracy result represents a single evaluation run over the corresponding evaluation set. For NLVR2 and some instances of COVR, the text input is a statement (to be classified as True/False). We convert each such statement to a question by adding the prefix ``Is it true that'' to it and converting the answer to ``yes''/``no.'' We use question embeddings to select 12 examples for GQA and 6 examples for COVR and NLVR2.

\section{Details on Baselines}
\label{app:baselines}
FewVLM randomly samples 16 few-shot examples for GQA. VisProg runs the program generation and execution pipeline five times, and for each iteration randomly samples 24 few-shot examples for GQA and 16 few-shot examples for NLVR2. ViperGPT uses 8 few-shot examples for GQA. VisProg uses \texttt{text-davinci-002} or \texttt{text-davinci-003} for code generation (according to the code release), while ViperGPT uses \texttt{code-davinci-002}.

\begin{figure}[ht]
    \vspace{-2mm}
    \includegraphics[width=0.48\textwidth]{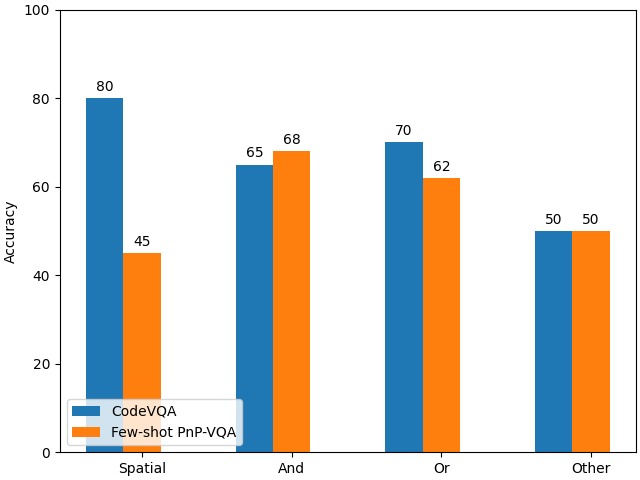}
    \caption{\textbf{Accuracy by question type in 2000 GQA validation examples.} \textcolor{tab-blue}{\texttt{CodeVQA} (blue)} outperforms \textcolor{tab-orange}{Few-shot PnP-VQA (orange)} on the spatial and or questions. ``Spatial'' refers to questions focusing on left/right or top/bottom relations or object positions.}
    \label{fig:gqa_breakdown_val}
    \vspace{-5mm}
\end{figure}

\begin{table*}[t]
\small
\centering
\captionsetup{font=footnotesize}
\resizebox{1.0\textwidth}{!}{
\begin{tabular}{l|ccc|cccc}
\toprule
\multirow{2}[3]{*}{{\bf Model}} & \multicolumn{3}{c}{GQA} & \multicolumn{4}{c}{COVR} \\
& \multicolumn{1}{c}{Shots} & \multicolumn{1}{c}{\textbf{Val Sample}} &  \multicolumn{1}{c}{\textbf{Testdev}} &  \multicolumn{1}{c}{Shots} & \multicolumn{1}{c}{\textbf{Val Sample}} & \multicolumn{1}{c}{\textbf{Val}} &  \multicolumn{1}{c}{\textbf{Test}} \\
\midrule
Few-shot PnP-VQA & 12 & 49.4 & 44.9 & 6 & 51.4 & -- & -- \\
\hspace{3mm}w/ \texttt{text-davinci-003} & & & & & & & \\
\texttt{CodeVQA} (ours) & 12 & 52.5 & 46.8 & 6 & 54.4 & -- & -- \\
\hspace{3mm}w/ \texttt{text-davinci-003} & & & & & & & \\
Few-shot PnP-VQA & 12 & 52.1 & 46.6 & 6 & 49.0 & 47.8 & 45.8 \\
\hspace{3mm}w/ \texttt{code-davinci-002} & & & & & & & \\
\texttt{CodeVQA} (ours) & 12 & \textbf{55.3} & \textbf{49.0} & 6 & \textbf{54.5} & \textbf{52.9} & \textbf{50.7} \\
\hspace{3mm}w/ \texttt{code-davinci-002} & & & & & & & \\
\bottomrule
\end{tabular}
}
\caption{\textbf{Validation and test results on GQA and COVR.} OpenAI model name (\texttt{text-davinci-003} or \texttt{code-davinci-002}) denotes which model was used as the question-answering model. GQA validation sample contains 2000 examples from the GQA validation set. COVR validation sample contains 1000 examples from the COVR non-paraphrased validation set. Highest scores on are in \textbf{bold}.}
\label{tab:results-with-val}
\end{table*}

\begin{figure*}[h]
    \centering
    \includegraphics[width=\textwidth]{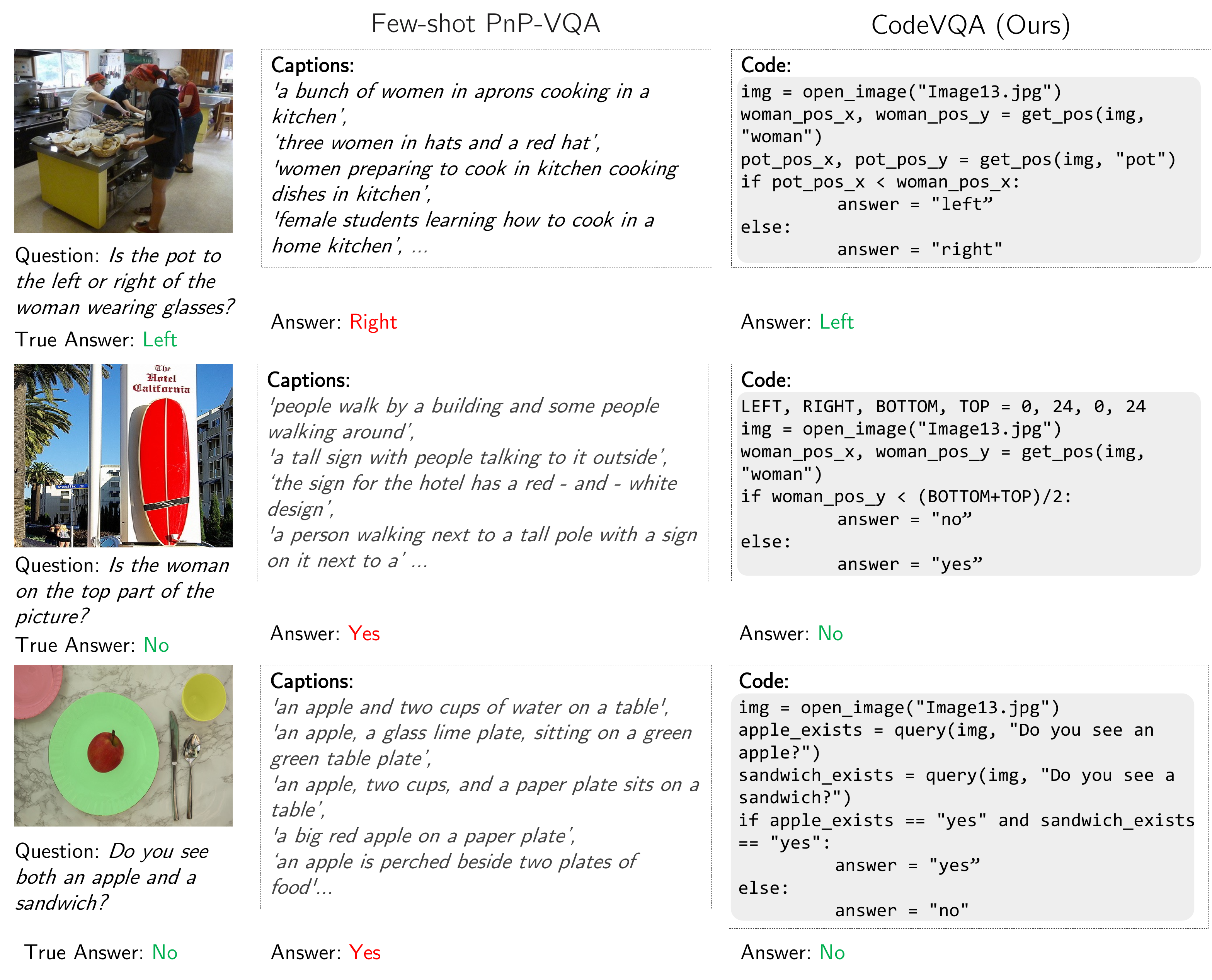}
    \caption{\small{\textbf{GQA Results}. We show example results from the GQA dataset where our method \texttt{CodeVQA} outperforms the baseline Few-Shot PnP-VQA. }
    }
    \label{fig:res2}
    \vspace{-5mm}
\end{figure*}

\section{Qualitative Comparisons}
\label{app:qualitative}
We include qualitative comparisons of our method CodeVQA to the baseline Few-shot PnP-VQA~(text-davinci-003) in Fig~\ref{fig:res2}. In all the instances, we can see that PnP-VQA produces captions that are irrelevant to the question, resulting in incorrect answers. On the other hand, \texttt{CodeVQA} breaks down the question into a Python code block. \texttt{CodeVQA} uses if-else conditions along with the pre-defined visual modules \texttt{get\_pos(image, text)} and \texttt{query(image, text)} to focus on the right regions of the image, arriving at the correct answer in an explainable fashion.

\begin{figure*}[h]
    \centering
    \includegraphics[width=\textwidth]{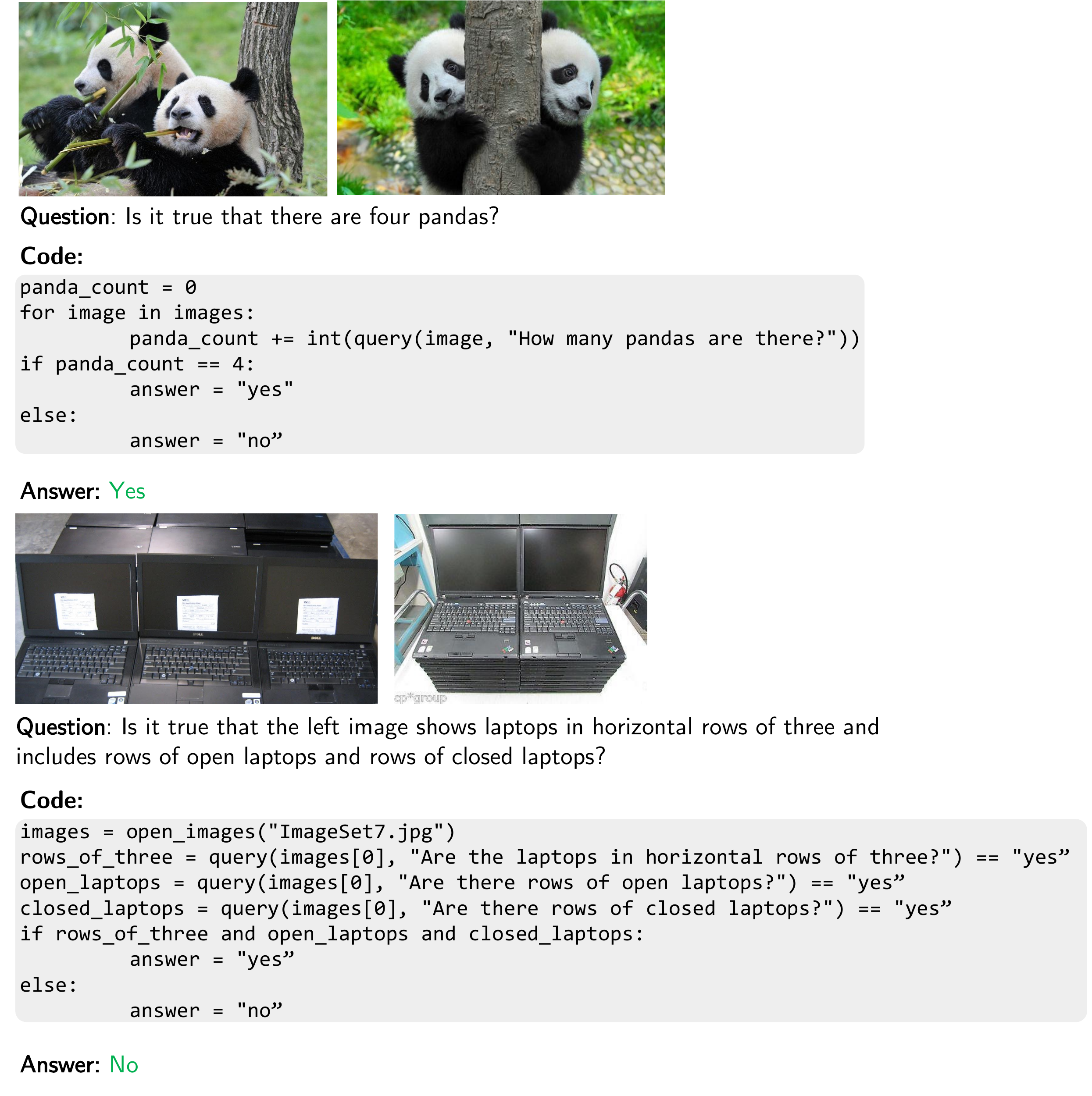}
    \caption{\small{\textbf{NLVR2 Results}. We show example results from the NLVR-2 dataset of our method \texttt{CodeVQA}.}
    }
    \label{fig:res3}
\end{figure*}

Fig.~\ref{fig:res3} shows two examples from the NLVR-2 dataset where our method \texttt{CodeVQA} answers the questions correctly. In the first example, it queries each of the images for the count of the pandas, and answers the question correctly based on that. In the second example, our method breaks the question down into three simpler queries and an \texttt{if-else} statement to arrive at the correct answer.  

\begin{figure*}
    \centering
    \includegraphics[width=1.2\textwidth]{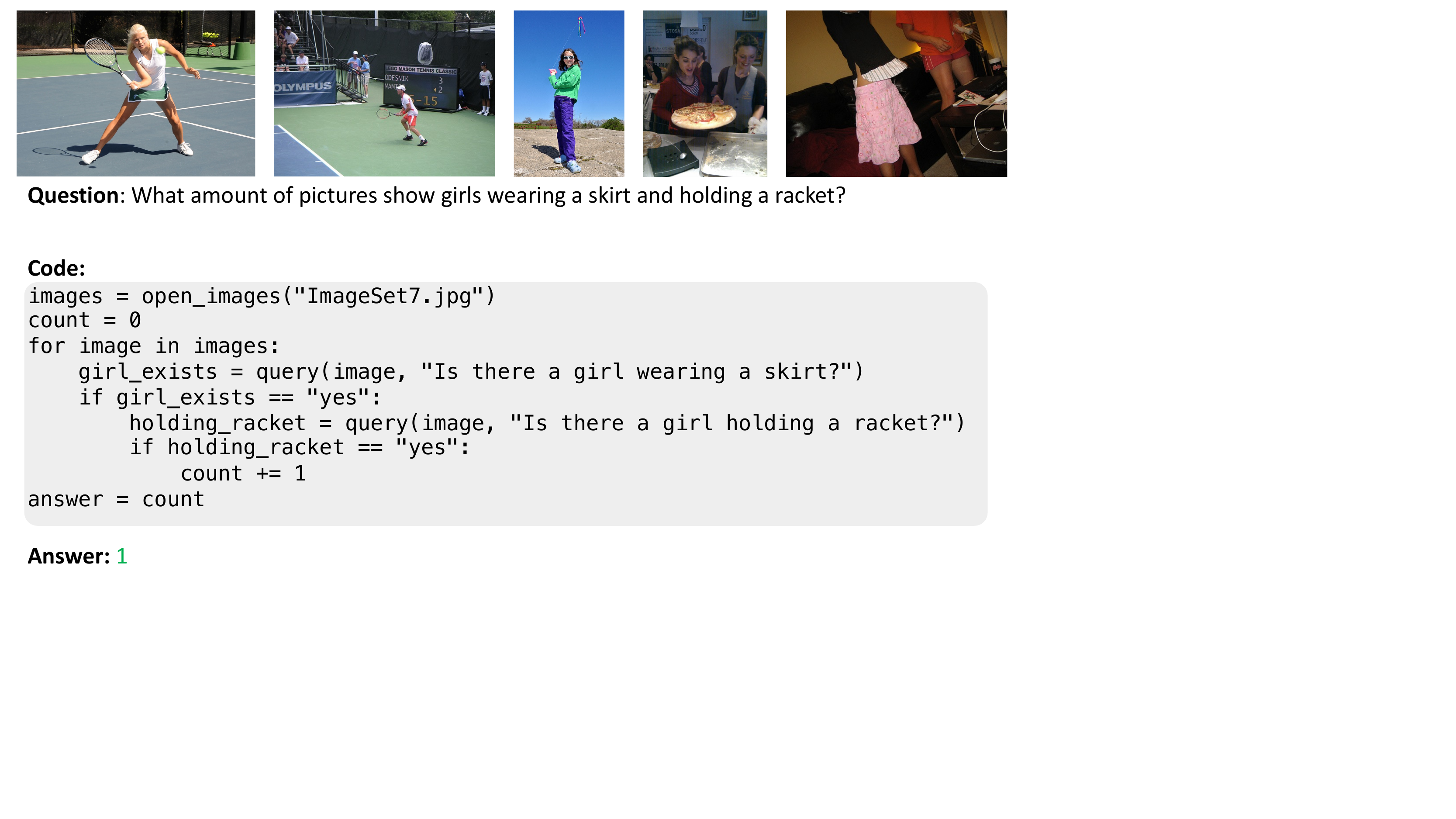}
    \includegraphics[width=1.2\textwidth]{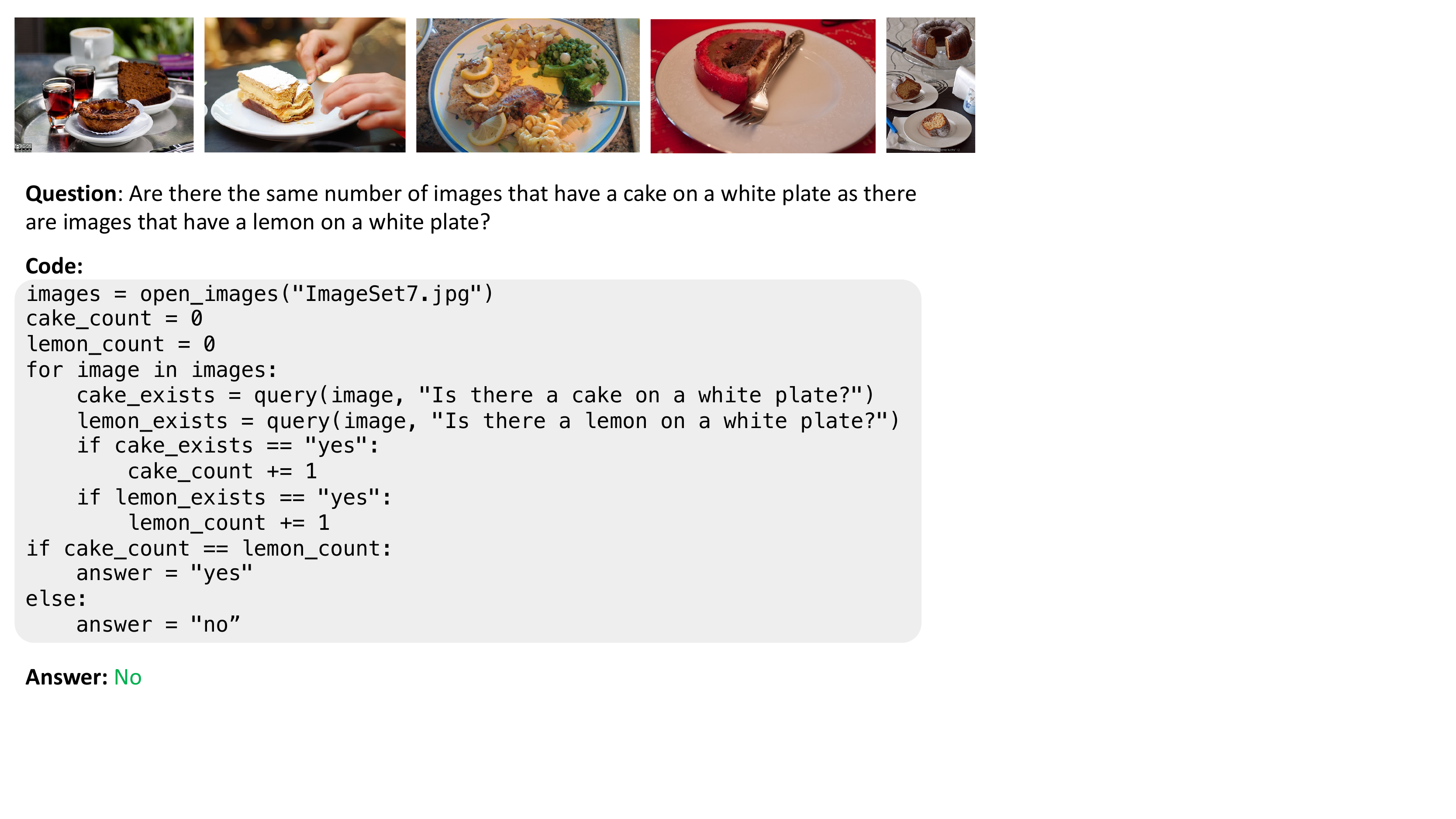}
    \caption{\small{\textbf{COVR Results}. We show results on the COVR dataset where our method correctly answers the question by referencing all the images.}    
    }
    \label{fig:res4}
\end{figure*}

Fig.~\ref{fig:res4} shows the correct results of our method on complex multireference questions in the COVR dataset. \texttt{CodeVQA} is able to break down the logic to obtain the counts of images with a cake on a white plate and images with a lemon on a white plate and then evaluates if the two counts are the same.  

In the second more complex example, our method uses \texttt{for} loops and complex \texttt{if-else} logic to first locate the images that satisfy the criterion, \emph{``pillows on a couch near a table''} and  \emph{``pillows on a couch near a bed''} to count the individual occurrences.   

\section{Additional Quantitative Results}
\label{app:quant-results}
Table~\ref{tab:results-with-val} shows results on validation sets and compares the accuracies of \texttt{CodeVQA} and Few-shot PnP-VQA when using \texttt{code-davinci-002} and \texttt{text-davinci-003} as the question-answering LM.

Table~\ref{tab:shot-number} shows how the accuracies of \texttt{CodeVQA} and Few-shot PnP-VQA vary with the number of shots in the prompt.
\begin{table}[ht]
\small
\centering
\captionsetup{font=footnotesize}
\resizebox{0.5\textwidth}{!}{
\begin{tabular}{lccc}
\toprule
\multirow{2}[3]{*}{{\bf Method}} & \multicolumn{3}{c}{Number of shots} \\
 & 8 & 12 & 16 \\
 \midrule
 \textit{text-davinci-003} & & & \\
Few-shot PnP-VQA & 48.3 & 49.4 & 49.5 \\
\texttt{CodeVQA} & 52.8 & 52.5 & 52.7 \\
\midrule
\textit{code-davinci-002} & & & \\
Few-shot PnP-VQA & 50.6 & 52.1 & 51.2 \\
\texttt{CodeVQA} & 55.1 & 55.3 & 55.4 \\
\bottomrule
\end{tabular}
}
\caption{Accuracy with different numbers of shots on 2000 GQA validation examples.}
\label{tab:shot-number}
\end{table}
Figure~\ref{fig:gqa_breakdown_val} shows the breakdown of accuracy by question type for 2000 GQA validation examples, which we used for initial experimentation (similar to Figure~\ref{fig:gqa_breakdown2} but on validation examples). We note that on this sample, Few-shot PnP-VQA has an advantage on ``and'' questions.

\section{Experiments with Additional Primitives}
\label{app:additional}
We also experiment with two other primitives, on datasets involving counting objects or knowledge retrieval:

\paragraph{\texttt{find\_object(image, object\_description)}} This function returns a set of references to objects in the image that match the given description, and we use it for counting objects. We implement this function using Grounding DINO \citep{liu2023grounding}, which is an open-vocabulary object detector that is also trained on referring expression comprehension.

\noindent
We evaluate this primitive on the VQAv2 dataset \citep{vqav2}, for which we use only this primitive and \texttt{query}, as well as the COVR and NLVR2 datasets. We used 12 in-context examples for the VQAv2 dataset. Table~\ref{tab:find-object} shows the results indicating that using this module for counting rather than \texttt{query} yields mixed results. Qualitatively, we observe a few reasons for errors in the \texttt{find\_object} version. First, the object detector is not always accurate (e.g. finding ``person holding a banana'' when there is a person but no banana). Second, our program may omit key details from the question (e.g. for ``How many boats have people in them?'' the program counts the number of boats overall). Third, our program may invoke the detector when it is ill-suited to the question (e.g. ``How many blades of grass surround the fire hydrant?''). On the other hand, captions often convey the number of objects when the number is small, which is very common in these datasets, so \texttt{query} can be effective on counting.

\paragraph{\texttt{knowledge\_query(question)}} This function returns the answer to a question based on world knowledge (e.g. ``Which football team has won the most Super Bowls?''). We implement this function using the same LM that is used for \texttt{query}. In order to better match the format of the OK-VQA dataset, we add a large negative bias to the logits of the following tokens to prevent the LM from generating them: hyphens, ``to'', and $^{\circ}$. This choice was made based on preliminary experiments on the OK-VQA dataset.

\noindent
We evaluate this primitive on the OK-VQA dataset \citep{okvqa}, for which we use only this primitive and \texttt{query}. For \texttt{CodeVQA} and Few-shot VQA, we used 7 in-context examples to be consistent with the OK-VQA results of ViperGPT \citep{vipergpt}. Table~\ref{tab:okvqa} provides the results, showing that for questions involving both visual information and general knowledge, breaking down the questions in this way does not lead to improved accuracy.

\noindent
For both VQAv2 and OK-VQA, we use the standard evaluation method associated with the VQAv2 dataset, which takes into account the set of ground-truth answers for each question. The Flamingo \citep{flamingo} results that we report on both datasets used 32 in-context examples.
\begin{table}[ht]
\small
\centering
\captionsetup{font=footnotesize}
\resizebox{0.5\textwidth}{!}{
\begin{tabular}{l|c|c|c}
\toprule
& VQAv2 & COVR & NLVR2 \\
 \midrule
 \textbf{Zero-shot} & & & \\
 BLIP-v2 & 65.0 & -- & -- \\
 \midrule
 \textbf{Few-shot} & & & \\
 Flamingo & 67.6\textdaggerdbl & -- & -- \\
Few-shot PnP-VQA & 66.84 & 47.8 & 63.4 \\
\texttt{CodeVQA} & -- & 52.9 & 64.0 \\
\texttt{CodeVQA} & 66.63 & 52.9 & 66.0 \\
\hspace{3mm}w/ \texttt{find\_object} & & & \\
\bottomrule
\end{tabular}
}
\caption{Results with \texttt{find\_object} used for counting objects on VQAv2 (a random sample of 4000 examples from validation set), COVR (validation), and NLVR2 (test-public). \textdaggerdbl indicates a result on the full VQAv2 test-dev set, which may not be directly comparable with our results on a sample of the validation set.}
\label{tab:find-object}
\end{table}

\begin{table}[ht]
\small
\centering
\captionsetup{font=footnotesize}
\resizebox{0.5\textwidth}{!}{
\begin{tabular}{l|c}
\toprule
& OK-VQA \\
 \midrule
 \textbf{Zero-shot} & \\
 BLIP-v2 & 45.9 \\
 \midrule
 \textbf{Few-shot} & \\
Flamingo & 57.8 \\
ViperGPT & 51.9 \\
Few-shot PnP-VQA & 54.1 \\ 
\texttt{CodeVQA} & 53.5 \\ 
\hspace{3mm}w/ \texttt{knowledge\_query} & \\
\bottomrule
\end{tabular}
}
\caption{Results with \texttt{knowledge\_query} on the OK-VQA validation set.}
\label{tab:okvqa}
\end{table}

\section{Licenses and Other Dataset Details}
\label{app:datasets}
GQA is licensed under the CC-BY-4.0 license (\href{https://creativecommons.org/licenses/by/4.0/}{https://creativecommons.org/licenses/by/4.0/}). The COVR repository (\href{https://github.com/benbogin/covr-dataset}{https://github.com/benbogin/covr-dataset}) is licensed under an MIT license (though imSitu images may not be licensed). The text in both datasets is written in English. The annotations in NLVR2 are licensed under CC-BY-4.0, but the images in the dataset are not licensed. The annotations in VQAv2 are licensed under CC-BY-4.0.

\noindent
The testdev set of GQA contains 12578 instances. The test set of COVR contains 7024 instances. The validation set of COVR contains 6891 instances. The public test set of NLVR2 contains 6967 instances. The validation set of OK-VQA contains 5046 instances. For VQAv2, we evaluate on a random sample of 4000 examples from the validation set.

\noindent
During the development and intermediate evaluations of our method, we evaluated on a random sample of 200 training examples and a random sample of 2000 validation examples from GQA, a random sample of 200 training examples and the validation set from COVR, a random sample of 2000 training examples from NLVR2, a random sample of 1200 training examples from OK-VQA, and a random sample of 2200 training examples from VQAv2.

\section{Ethics and Impact Statement}
One goal of our work is to decrease the need for (re-)training VQA systems. Achieving this goal would mean a decrease in carbon emissions from training models. However, our approach also has a high inference cost, given the use of large language models. A decision to employ our approach should take into consideration this computational cost and the associated environmental impact.

Another potential positive impact of our approach is improved interpretability via the generated programs. These programs offer to people familiar with Python a record of which visual tasks the system uses for a given question and how the system combines the outputs of these tasks to predict the answer.

Our system relies on pre-trained vision-language models to predict answers to visual questions. Prior work \citep{Ross2020MeasuringSB,Agarwal2021EvaluatingCT} has found evidence of social biases in vision-language models trained on image-captions. Therefore, our system may exhibit these biases as well. Practitioners should be aware of this risk and ideally should take steps to mitigate this risk when they consider deploying this system in ways that can impact human lives.

\end{document}